\documentclass[pmlr,twocolumn,10pt]{jmlr}

\usepackage{comment}
\usepackage{booktabs}
\usepackage{siunitx}
\usepackage{tabularray}
\usepackage{colortbl}
\UseTblrLibrary{booktabs}

\theorembodyfont{\upshape}
\theoremheaderfont{\scshape}
\theorempostheader{:}
\theoremsep{\newline}

\jmlrvolume{LEAVE UNSET}
\jmlryear{2023}
\jmlrsubmitted{LEAVE UNSET}
\jmlrpublished{LEAVE UNSET}
\jmlrworkshop{Machine Learning for Health (ML4H) 2023} 

\title[KG Representations to enhance ICU Time-Series Predictions]{Knowledge Graph
Representations to enhance \\ Intensive Care
Time-Series Predictions}

\author{\Name{Samyak Jain} \Email{jainsamyak2512@gmail.com} \\
\Name{Manuel Burger} \Email{manuel.burger@inf.ethz.ch} \\
\Name{Gunnar Rätsch} \Email{raetsch@inf.ethz.ch} \\
\Name{Rita Kuznetsova} \Email{rita.kuznetsova@inf.ethz.ch} \\
\addr Department of Computer Science, ETH Zürich, Switzerland
}

\begin{document}

\maketitle

\begin{abstract}
Intensive Care Units (ICU) require comprehensive patient data integration for enhanced clinical outcome predictions, crucial for assessing patient conditions. Recent deep learning advances have utilized patient time series data, and fusion models have incorporated unstructured clinical reports, improving predictive performance. However, integrating established medical knowledge into these models has not yet been explored. The medical domain's data, rich in structural relationships, can be harnessed through knowledge graphs derived from clinical ontologies like the Unified Medical Language System (UMLS) for better predictions. Our proposed methodology integrates this knowledge with ICU data, improving clinical decision modeling. It combines graph representations with vital signs and clinical reports, enhancing performance, especially when data is missing. Additionally, our model includes an interpretability component to understand how knowledge graph nodes affect predictions.

\end{abstract}

\section{Introduction}
\label{intro}
Recently, researchers have investigated the use of multimodal data types~\citep{khadanga2019using, husmann2022importance}, including patient time series and unstructured clinical reports, for predictive tasks in Intensive Care Units (ICU)~\citep{harutyunyan2019multitask, yeche2021hirid}. However, these approaches often overlook the rich structured relationships inherent in the medical domain, such as those in clinical ontologies like the Unified Medical Language System (UMLS)~\citep{bodenreider2004unified}. These ontologies exhibit a structural prior, which can be exploited by appropriate modeling architectures to capture the complex interdependencies of medical data better, ultimately leading to more accurate and reliable clinical predictions. To bridge this gap, we propose a methodology that integrates these structured relationships into ICU time series predictions using knowledge graphs (KGs). Our approach, which involves jointly learning graph representations alongside time series and clinical reports, significantly enhances model performance, particularly in cases with missing data. The main contributions are as follows:
\begin{itemize}
    \item \textbf{Usage of prior medical knowledge}: We construct KGs using the UMLS database, demonstrating that the incorporation of a strong structural prior into our model leads to enhanced performance.
    \item \textbf{Performance under Missing Data}: Graph structure-based methods have shown strong performance in scenarios with missing data~\citep{zhang2021graph}. Building on this, our approach diverges by jointly learning representations across various modalities, which further improves performance in situations with missing data.
    \item \textbf{Interpretability}: In the medical domain, machine learning systems must be not only reliable but also interpretable and explainable. We identify the most influential KG nodes and examine their impact on the model's predictions.
\end{itemize}

\begin{figure*}[tb]
    \centering
    \includegraphics[width=.9\linewidth]{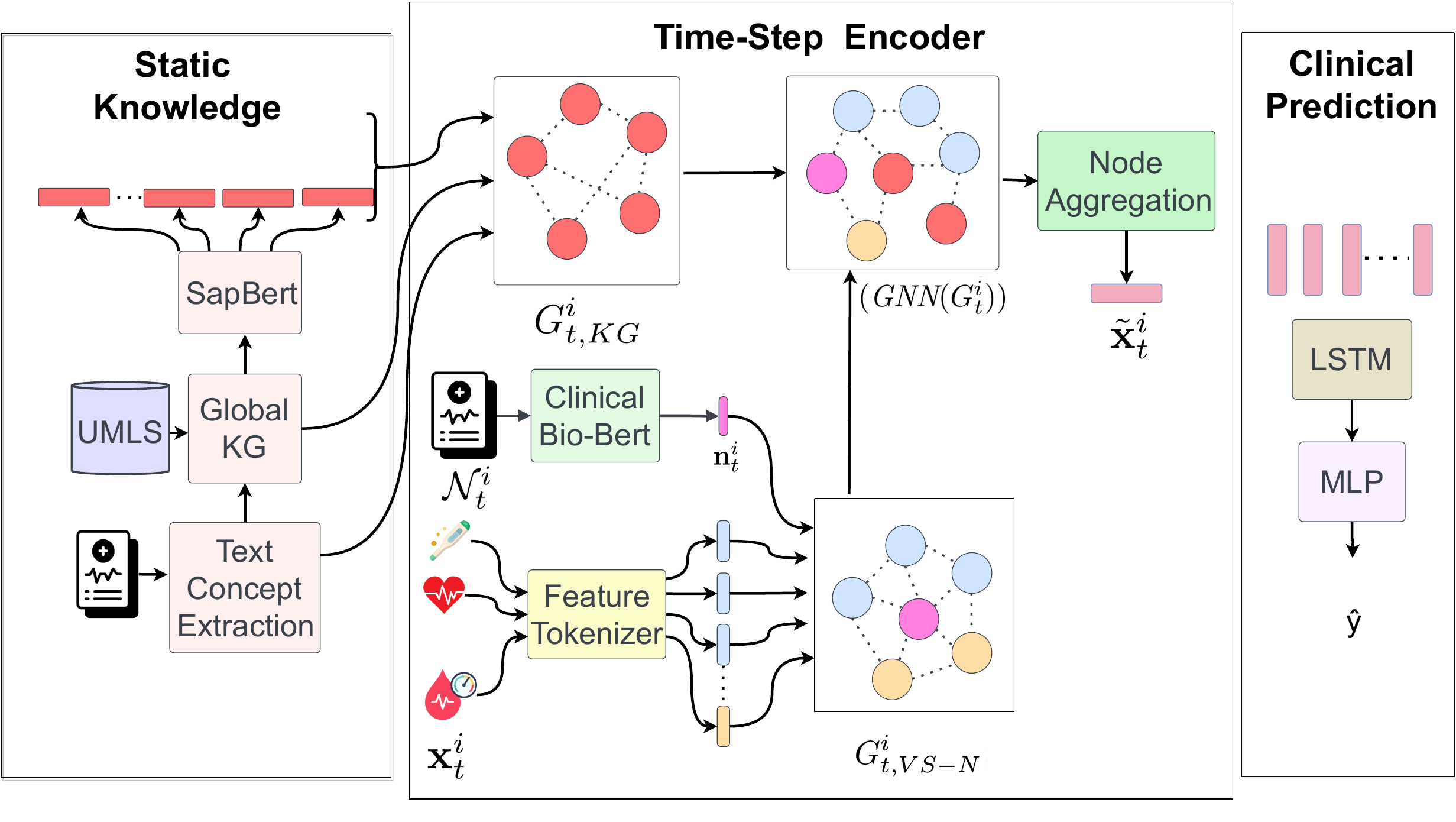}
    \caption{Architecture Diagram: We extract \emph{Static Knowledge} (Sec.~\ref{sec:static-knowledge}), we compute a rich multi-modal \emph{time-step embedding} including structured knowledge (Sec. \ref{sec:time-step-encoder}) and then perform clinical prediction tasks (Sec. \ref{sec:clinical-prediction}).}
    \label{fig:arch}
\end{figure*}
Experimental results from the MIMIC-III Benchmark tasks~\citep{johnson2016data, harutyunyan2019multitask} show that the use of prior knowledge not only enhances predictive performance but also yields better results with increasing amounts of missing data. Additionally, our interpretability analysis highlights the most predictive medical concepts within our knowledge graph.

\section{Related Work}
\label{related}
In the subsequent sections, we discuss related work on multi-modal approaches, encompassing Large Language Models and the use of prior knowledge through KGs, in the context of ICU data.

\paragraph{Multi-Modality} Prior research in learning representations from multimodal data has explored a wide range of approaches. These approaches include: combining sequence models with convolutional feature extractors to process clinical notes~\citep{khadanga2019using}; integrating sequence models with BERT~\citep{devlin2018bert, deznabi2021predicting}; leveraging Named Entity Recognition models in conjunction with word and document embeddings to capture structured information~\citep{bardak2021improving} and employing cross-modal transformers to effectively model relationships between different modalities~\citep{husmann2022importance}.

\paragraph{Graph Approaches} Recent studies have explored the utilization of relational structures among different variables through the application of Graph Neural Networks (GNNs). \citet{zhang2021graph} introduced a model that leverages these relational structures. Another approach~\citep{zhang2022pm2f2n} incorporates a patient correlation graph fusion model. However, these works do not incorporate prior medical knowledge from diverse sources and vocabularies and are also not suitable for online prediction tasks. Alternatively, there is a separate line of research that utilizes such knowledge in the form of knowledge graphs to enhance predictions. Notably, these knowledge-enhanced approaches are primarily focused on tasks related to subsequent hospital admissions~\citep{burger2023multi, jiang2023graphcare} or non-sequence tasks~\citep{roy2021incorporating, chandak2023building} rather than multi-variate ICU time series analysis.

\paragraph{Language Models} Finally, language models have been proposed to solve clinical tasks~\citep{dragon-yasunaga2022deep, gpt4_medical-nori2023capabilities}. However, such models cannot natively handle the diverse set of numerical and categorical clinical variables in the form of a multi-variate time series to perform prediction tasks in an intensive care setting.

\paragraph{Our Work} In this study, we leverage prior medical knowledge, clinical notes, and graph structures to enhance multi-variate ICU time series models. Our approach also explores how the combination of these modalities can mitigate challenges in scenarios with missing data. These models are designed with a specific focus on ICU prediction tasks while remaining suitable for online predictions (contiguous predictions at every time step).

\begin{table*}[t]
    \small
    \centering
    \caption{Model Performance Comparison on MIMIC-III Benchmark tasks~\citep{harutyunyan2019multitask}.}
    \begin{tabular}{l l l l l l l}
    \toprule
    \bf Method & \multicolumn{2}{c}{\textbf{Mortality}} & \multicolumn{2}{c}{\textbf{Decompensation}} & \multicolumn{2}{c}{\textbf{Phenotyping}} \\
      & \textit{AuPRC} & \textit{AuROC} & \textit{AuPRC} & \textit{AuROC} &  \textit{macro-AUC} & \textit{micro-AUC} \\
    \midrule

\cite{harutyunyan2019multitask}  & 50.1 $\pm$ 1.3 & 86.1 $\pm$ 0.3 & 34.1 $\pm$ 0.5 & 90.7 $\pm$ 0.2 & 77.6 $\pm$ 0.4  & 82.5 $\pm$ 0.3\\
    \cite{khadanga2019using} & 52.5 $\pm$ 1.3 & 86.5 $\pm$ 0.4 & 34.5 $\pm$ 0.7 & 90.7 $\pm$ 0.7 & - & - \\
    \cite{husmann2022importance} & 52.7 $\pm$ 1.0 & 87.1 $\pm$ 0.6 & 39.7 $\pm$ 0.6 & 92.2 $\pm$ 0.2 &  82.6 $\pm$ 0.1 & 86.1 $\pm$ 0.1\\
    \cite{yang2021multimodal} & 56.2 $\pm$ na & 85.7 $\pm$ na & - & - & - & - \\
\cite{zhang2021graph} & 44.6 $\pm$ 2.1 & 83.7 $\pm$ 0.5 & - & - & - & -\\
    \emph{Ours} & \textbf{58.7 $\pm$ 0.6} & \textbf{89.3 $\pm$ 0.1} & \textbf{47.2 $\pm$ 1.0} & \textbf{94.2 $\pm$ 0.3} & \textbf{84.6 $\pm$ 0.1} & \textbf{87.9 $\pm$ 0.1} \\

    \bottomrule
    \end{tabular}
    \label{tab:performance-comparison}
\end{table*}

\section{Methodology}
\label{methods}

In this section, we formally introduce our method and how we incorporate prior knowledge into the time-series model. We consider a dataset $\mathcal{D}$ of multiple patient time series. The time-series of patient $i$ of length $T$ is $\mathbf{X}^{i} = [\mathbf{x}_{1},\mathbf{x}_{2},\dots,\mathbf{x}_{T}]$ and consists of time-steps $\mathbf{x}_{t}^{i}=[x_{t,1},x_{t,2},\dots,x_{t,n_{VS}}] \in \mathbb{R}^{n_{VS}}$, where we observe $n_{VS}$ vital signs measurements. Additionally, let $\mathcal{N}_{t}^{i}$ be the set of clinical notes available at time $t$ for patient $i$ (details on clinical notes in App.~\ref{app:clinical-notes}). We obtain the aggregated representation of all clinical notes at each time-step $\mathbf{n}_{t}^{i} = \texttt{BERT}(\mathcal{N}_{t}^{i}) \in \mathbb{R}^{d}$ using a pre-trained Clinical Bio-Bert model~\citep{alsentzer2019publicly}.

\subsection{Static Knowledge}
\label{sec:static-knowledge}

We build a concept extraction pipeline using QuickUMLS~\citep{soldaini2016quickumls} to obtain the set of medical concepts found in clinical notes $V_{t}^{i} = \texttt{QuickUMLS}(\mathcal{N}_{t}^{i})$. We then build a global medical knowledge graph $G = (V, E)$ as follows in Eqn.~\ref{eqn:global-kg}:
\begin{equation}
\label{eqn:global-kg}
    \begin{gathered}
        V = \bigcup_{i,t \in \mathcal{D} \times T} V_{t}^{i},\\
     E = \{ \{u, v\}  \,|\, \forall \{u, v\} \in UMLS \wedge u,v \in V\}
    \end{gathered}
\end{equation}
with nodes $V$ and (undirected) edges (as node sets $\{u, v\}$) from the UMLS database $E$. We use the SapBERT model~\citep{pubmedbert} on the node descriptors available in UMLS to obtain node representations. Figure~\ref{fig:arch} (left) demonstrates the process.

\subsection{Time-Step Encoder}
\label{sec:time-step-encoder}
In this section, we describe how to obtain a rich step embedding using different modalities for each patient $i$ at each time-step $t$. $\mathbf{x}_{t}^{i} \in \mathbb{R}^{n_{VS}}$ is given as input to the Feature Tokenizer (FT) module~\citep{gorishniy2021revisiting} to convert each of the $n_{VS}$ measurements into an embedding. We concatenate the clinical note representation $\mathbf{n}_{t}^{i}$ and obtain for each time-step the vital signs and text (clinical notes) matrix:
\begin{equation}
    \mathbf{X}_{t, VS-N}^{i} = [\texttt{FT}(\mathbf{x}_{t}^{i}); \mathbf{n}_{t}^{i}] \in \mathbb{R}^{(n_{VS}+1)\times d}
\end{equation}
We then build the vital signs and text (clinical notes) graph:
\begin{equation}
    G_{t, VS-N}^{i} = (\mathbf{X}_{t, VS-N}^{i}, E_{VS-N})
\end{equation}
where $E_{VS-N}$ corresponds to edge connectivity based on prior knowledge (e.g. organ system assignments of vitals) or full connectivity amongst vitals. The text node $\mathbf{n}_{t}^{i}$ is connected to all vitals.
To incorporate prior knowledge a subgraph $G_{t, KG}^{i}$ is formed by querying the global graph $G$ using the extracted concepts $V_{t}^{i}$ at the timepoint $t$ for patient $i$.
\begin{align}
    E_{t, KG}^{i} &= \{\{u, v\} \,|\, \forall\{u, v\} \in E \wedge u, v \in V_{t}^{i} \} \\
    G_{t, KG}^{i} &= (\mathbf{X}_{t, KG}^{i}, E_{t, KG}^{i})
\end{align}

where $\mathbf{X}_{t, KG}^{i}$ are the SapBERT node embeddings and $E_{t, KG}^{i}$ are edges from the global knowledge graph $G$ (see Sec. \ref{sec:static-knowledge}).
We then join $G_{t, VS-N}^{i}$ and $G_{t, KG}^{i}$ by adding undirected edges from all nodes in $G_{t, VS-N}^{i}$ to all nodes in $G_{t, KG}^{i}$ to obtain the final dynamic time-step graph $G_{t}^{i}$:
\begin{align}
    \overline{E}_{t}^{i} &= \{\{u, v\} \,|\, \forall u \in G_{t, VS-N}^{i},\, \forall v \in G_{t, KG}^{i} \} \\
    E_{t}^{i} &= \overline{E}_{t}^{i} \cup E_{t, KG}^{i} \cup E_{VS-N} \\
    G_{t}^{i} &= ( \mathbf{X}_{t, VS-N}^{i} \cup \mathbf{X}_{t, KG}^{i},\, E_{t}^{i} )
\end{align}

Information is exchanged among the nodes of each dynamic graph $G_{t}^{i}$ using GNN layers and finally, all nodes are aggregated to form a rich \emph{step embedding} $\Tilde{\mathbf{x}}_{t}^{i} = \texttt{AGG}( \textit{GNN}(G_{t}^{i})) \in \mathbb{R}^{d}$. For \texttt{AGG} we use non-parametric aggregation functions such as $sum$, $mean$, or $max$. Figure~\ref{fig:arch} (middle) demonstrates the process.

\subsection{Clinical Prediction}
\label{sec:clinical-prediction}
Based on the previous Section \ref{sec:time-step-encoder} we obtain a \emph{step embedding} $\Tilde{\mathbf{x}}_{t}^{i}$ at each time point. The sequence of time step embeddings is processed by a sequence model (such as LSTM~\citep{lstm-hochreiter-1997}) and followed by a Multi-Layered Perceptron (MLP) that uses the hidden state(s) of the LSTM network to generate the final prediction.

\section{Experimental Setup}
\label{experiments}
\paragraph{Data Processing}
We perform our experiments on the MIMIC-III dataset~\citep{johnson2016data}. To avoid data leakage we remove all discharge summary notes and mask out the last available notes as they often contain explicit mortality terms~\citep{husmann2022importance}. For more details, we refer to the App. \ref{sec:data}.   

\paragraph{Tasks}
We analyze the performance of the model on established benchmark tasks~\citep{harutyunyan2019multitask}. We focus on 3 clinical prediction tasks. 1)~In-hospital mortality is a binary prediction task using data from the first 48 hours of patient stay. 2)~Decompensation is an hourly online prediction task with a binary label stating whether a patient will die in the next 24 hours or not. 3)~Phenotyping uses the entire time series of a patient stay and classifies it into 25 acute care conditions. 

\paragraph{Architecture}
Different hyperparameters of our architecture were selected using a grid/random search approach to optimize validation set performance on the mortality task. Details can be found in App.~\ref{sec:hyper}.

\section{Results}
\label{results}
We compare the performance of the proposed solution with other existing multi-modal fusion methods. Table \ref{tab:performance-comparison} shows the comparison and it is evident that we outperform other methods. \citet{harutyunyan2019multitask} provided initial benchmark baselines using different time-series architectures on vital signs data. \cite{khadanga2019using, yang2021multimodal} propose multi-modal fusion architectures with separate encoders for each modality and \citet{husmann2022importance} benchmark a multi-modal transformer with cross-attention across the modalities at every layer. Finally, \citet{zhang2021graph} introduced a graph-structured model with increased robustness on missing data to process irregularly sampled vital signs time series. Our results highlight the benefits of pairing multi-modal fusion with prior knowledge using graph-structured time-step embeddings.

\begin{figure}[t]
    \centering
    \includegraphics[width=0.9\linewidth]{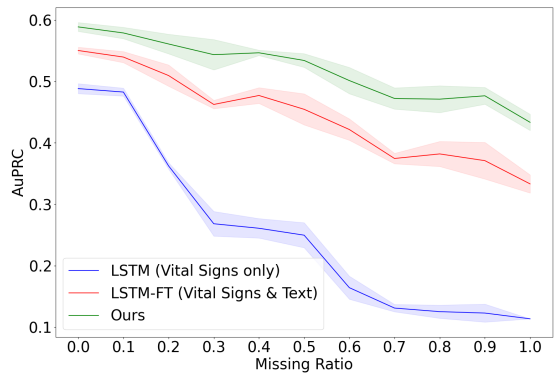}
    \caption{Performance under Missing Data on the Mortality prediction task after 48 hours of patient stay. Shaded areas indicate standard deviations.}
    \label{fig:missing}
\vspace{-1em}
\end{figure}

\begin{figure}
    \centering
    \includegraphics[width=.75\linewidth]{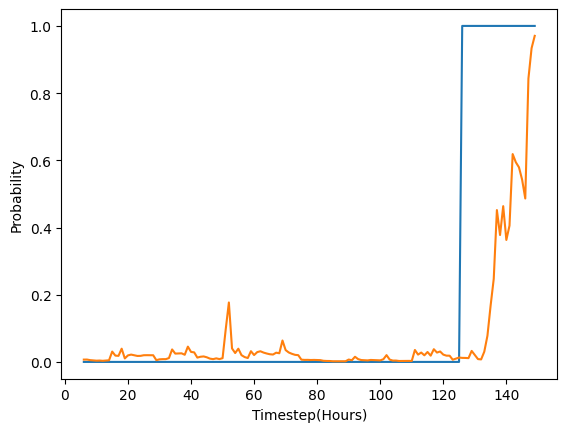}
    \caption{Decompensation Analysis of Patient ID \texttt{11486}: \textit{orange} line shows the model's predicted probability for the binary decompensation label and the \textit{blue} line depicts the ground-truth label}
    \label{fig:decompensation-score-plot}
    \vspace{-1em}
\end{figure}

\begin{figure*}[h!]
    \centering
        \begin{minipage}{.5\textwidth}
        \centering
        \includegraphics[width=.85\linewidth]{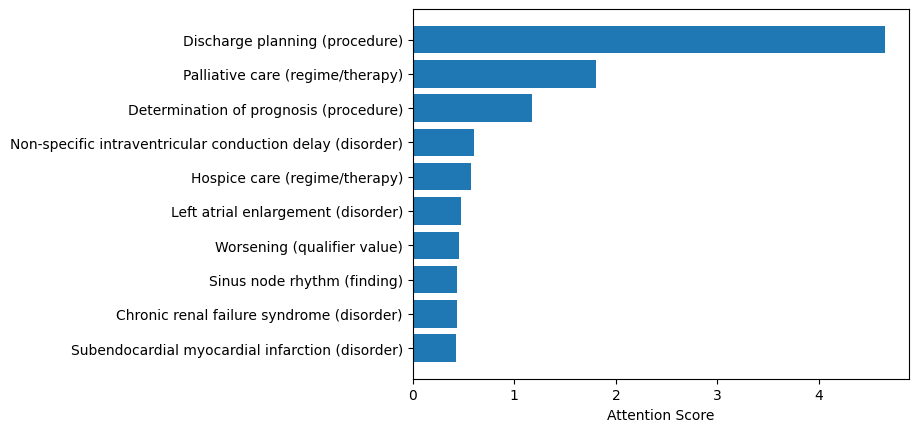}
        \end{minipage}\begin{minipage}{.5\textwidth}
        \centering
        \includegraphics[width=1.3\linewidth]{images/ex-1h.pdf}
        \vspace{0.1cm}
        \end{minipage}
    \caption{Decompensation Analysis of Patient ID \texttt{11486} : (left) Attention Scores of top 10 KG nodes aggregated the entire time-series; (right) Attention Score HeatMap(s) at timestep 138 of the two employed GNN layers. Lower indices correspond to vital signs nodes from $G_{t, VS-N}^{i}$ and higher indices correspond to nodes from the time-point specific (sub-)knowledge graph $G_{t, KG}^{i}$.}
    \label{fig:decompensation-attention-plot}
    \vspace{-0.5em}
\end{figure*}

\paragraph{Performance under Missing Data}
We ascertain the model's performance under missing data by randomly masking vital signs data of the test set~\citep{zhang2021graph}. Figure \ref{fig:missing} shows how the performance of different methods varies with different missing ratios. A missing ratio refers to the amount of masking e.g. $missing\;ratio=0.5$ implies 50\% of the vital signs data is masked. We observe increased robustness under missing data in case of multi-modal inputs (e.g. including text, red curve) and an additional improvement under our proposed rich, multi-modal, knowledge-enhanced, and graph-structured step embedding (green curve).

We do note, that models considering multiple modalities such as text and/or knowledge graph concepts perform well even under high amounts of missing vital signs data. Inspection of relevant clinical notes reveals explicit discussion of hospital staff on discharge planning or certain disorders, which are highly indicative for tasks such as \emph{mortality} or \emph{decompensation}. This behavior is further emphasized in our proposed method by explicitly extracting related medical concepts and not only embedding entire pieces of text. In the following paragraph, we show strong attention of the model on task-related knowledge graph nodes such as \emph{palliative care}.

\paragraph{Interpretability}

Leveraging a Graph Attention Network (GAT) by~\citet{velivckovic2017graph}, we compute attention scores across KG nodes, highlighting the significance of procedural (e.g., discharge planning), therapeutic (e.g., palliative care and hospice care), and disorder-related nodes (see Figure~\ref{fig:decompensation-attention-plot} (left)).

We then conduct an interpretability analysis to investigate the model's predictions, emphasizing temporal dynamics as illustrated in Figure \ref{fig:decompensation-score-plot}. We can see a strong increase in the predicted probability of decompensation around time-point 140.  A focused examination of the time-step graph $G_{t}^{i}$ at time-step 138 provides insight into notable changes in the predicted probability of decompensation. Figure \ref{fig:decompensation-attention-plot} (right) shows the heatmaps after the first and second GNN layers at time step 138. They depict the attention matrix formed using different nodes of the time-step graph $G_{t}^{i}$, where lower indices correspond to nodes from $G_{t, VS-N}^{i}$ and higher indices correspond to nodes from $G_{t, KG}^{i}$ (Sec. \ref{sec:time-step-encoder}). From the left heatmap (i.e. the first GNN layer) the 24th entry has particularly high attention. The specific node is the knowledge graph concept corresponding to \emph{palliative care}. As \emph{palliative care} is specialized care for people with serious illnesses, the model appropriately learned its correlation with the decompensation target.

We conclude that the extraction of prior medical knowledge aids interpretability at a detailed level of individual time steps. Further, we would like to underline the relevance of extracting and structuring model inputs based on prior knowledge concepts to provide more precise signals to our learning algorithms.

\section{Conclusion}
\label{conclusion}

We investigated the impact of incorporating Knowledge Graphs, derived from prior medical knowledge, on ICU time series clinical prediction tasks. We introduced a novel method for extracting and utilizing this knowledge in ICU predictions. Our findings highlight the effectiveness of graph-structured architectures with prior knowledge, particularly in scenarios with increasing missing data, which is a common challenge in clinical data applications. Furthermore, we emphasize the significance of discretely extracted prior knowledge concepts, which enhance both interpretability and learning signals.

\acks{This project was supported by grant \#2022-278 of the Strategic Focus Area “Personalized Health and Related Technologies (PHRT)” of the ETH Domain (Swiss Federal Institutes of Technology).}

\nocite{icons}
\bibliography{bibliography}

\begin{thebibliography}{27}
\providecommand{\natexlab}[1]{#1}
\providecommand{\url}[1]{\texttt{#1}}
\expandafter\ifx\csname urlstyle\endcsname\relax
  \providecommand{\doi}[1]{doi: #1}\else
  \providecommand{\doi}{doi: \begingroup \urlstyle{rm}\Url}\fi

\bibitem[Alsentzer et~al.(2019)Alsentzer, Murphy, Boag, Weng, Jin, Naumann, and
  McDermott]{alsentzer2019publicly}
Emily Alsentzer, John~R Murphy, Willie Boag, Wei-Hung Weng, Di~Jin, Tristan
  Naumann, and Matthew McDermott.
\newblock Publicly available clinical bert embeddings.
\newblock \emph{arXiv preprint arXiv:1904.03323}, 2019.

\bibitem[Bardak and Tan(2021)]{bardak2021improving}
Batuhan Bardak and Mehmet Tan.
\newblock Improving clinical outcome predictions using convolution over medical
  entities with multimodal learning.
\newblock \emph{Artificial Intelligence in Medicine}, 117:\penalty0 102112,
  2021.

\bibitem[Bodenreider(2004)]{bodenreider2004unified}
Olivier Bodenreider.
\newblock The unified medical language system (umls): integrating biomedical
  terminology.
\newblock \emph{Nucleic acids research}, 32\penalty0 (suppl\_1):\penalty0
  D267--D270, 2004.

\bibitem[Burger et~al.(2023)Burger, R{\"a}tsch, and
  Kuznetsova]{burger2023multi}
Manuel Burger, Gunnar R{\"a}tsch, and Rita Kuznetsova.
\newblock Multi-modal graph learning over umls knowledge graphs.
\newblock \emph{arXiv preprint arXiv:2307.04461}, 2023.

\bibitem[Chandak et~al.(2023)Chandak, Huang, and Zitnik]{chandak2023building}
Payal Chandak, Kexin Huang, and Marinka Zitnik.
\newblock Building a knowledge graph to enable precision medicine.
\newblock \emph{Scientific Data}, 10\penalty0 (1):\penalty0 67, 2023.

\bibitem[Devlin et~al.(2018)Devlin, Chang, Lee, and Toutanova]{devlin2018bert}
Jacob Devlin, Ming-Wei Chang, Kenton Lee, and Kristina Toutanova.
\newblock Bert: Pre-training of deep bidirectional transformers for language
  understanding.
\newblock \emph{arXiv preprint arXiv:1810.04805}, 2018.

\bibitem[Deznabi et~al.(2021)Deznabi, Iyyer, and
  Fiterau]{deznabi2021predicting}
Iman Deznabi, Mohit Iyyer, and Madalina Fiterau.
\newblock Predicting in-hospital mortality by combining clinical notes with
  time-series data.
\newblock In \emph{Findings of the association for computational linguistics:
  ACL-IJCNLP 2021}, pages 4026--4031, 2021.

\bibitem[Gorishniy et~al.(2021)Gorishniy, Rubachev, Khrulkov, and
  Babenko]{gorishniy2021revisiting}
Yury Gorishniy, Ivan Rubachev, Valentin Khrulkov, and Artem Babenko.
\newblock Revisiting deep learning models for tabular data.
\newblock \emph{Advances in Neural Information Processing Systems},
  34:\penalty0 18932--18943, 2021.

\bibitem[Gu et~al.(2020)Gu, Tinn, Cheng, Lucas, Usuyama, Liu, Naumann, Gao, and
  Poon]{pubmedbert}
Yu~Gu, Robert Tinn, Hao Cheng, Michael Lucas, Naoto Usuyama, Xiaodong Liu,
  Tristan Naumann, Jianfeng Gao, and Hoifung Poon.
\newblock Domain-specific language model pretraining for biomedical natural
  language processing, 2020.

\bibitem[Harutyunyan et~al.(2019)Harutyunyan, Khachatrian, Kale, Ver~Steeg, and
  Galstyan]{harutyunyan2019multitask}
Hrayr Harutyunyan, Hrant Khachatrian, David~C Kale, Greg Ver~Steeg, and Aram
  Galstyan.
\newblock Multitask learning and benchmarking with clinical time series data.
\newblock \emph{Scientific data}, 6\penalty0 (1):\penalty0 96, 2019.

\bibitem[Hochreiter and Schmidhuber(1997)]{lstm-hochreiter-1997}
Sepp Hochreiter and J\"{u}rgen Schmidhuber.
\newblock Long short-term memory.
\newblock \emph{Neural Comput.}, 9\penalty0 (8):\penalty0 1735–1780, nov
  1997.
\newblock ISSN 0899-7667.
\newblock \doi{10.1162/neco.1997.9.8.1735}.
\newblock URL \url{https://doi.org/10.1162/neco.1997.9.8.1735}.

\bibitem[Husmann et~al.(2022)Husmann, Y{\`e}che, R{\"a}tsch, and
  Kuznetsova]{husmann2022importance}
Severin Husmann, Hugo Y{\`e}che, Gunnar R{\"a}tsch, and Rita Kuznetsova.
\newblock On the importance of clinical notes in multi-modal learning for ehr
  data.
\newblock \emph{arXiv preprint arXiv:2212.03044}, 2022.

\bibitem[Jiang et~al.(2023)Jiang, Xiao, Cross, and Sun]{jiang2023graphcare}
Pengcheng Jiang, Cao Xiao, Adam Cross, and Jimeng Sun.
\newblock Graphcare: Enhancing healthcare predictions with open-world
  personalized knowledge graphs.
\newblock \emph{arXiv preprint arXiv:2305.12788}, 2023.

\bibitem[Johnson et~al.(2016{\natexlab{a}})Johnson, Pollard, Shen, Lehman,
  Feng, Ghassemi, Moody, Szolovits, Anthony~Celi, and Mark]{johnson2016mimic}
Alistair~EW Johnson, Tom~J Pollard, Lu~Shen, Li-wei~H Lehman, Mengling Feng,
  Mohammad Ghassemi, Benjamin Moody, Peter Szolovits, Leo Anthony~Celi, and
  Roger~G Mark.
\newblock Mimic-iii, a freely accessible critical care database.
\newblock \emph{Scientific data}, 3\penalty0 (1):\penalty0 1--9,
  2016{\natexlab{a}}.

\bibitem[Johnson et~al.(2016{\natexlab{b}})Johnson, Pollard, Shen, Lehman,
  Feng, Ghassemi, Moody, Szolovits, Celi, and Mark]{johnson2016data}
Alistair~EW Johnson, Tom~J Pollard, Lu~Shen, Li-Wei~H Lehman, Mengling Feng,
  Mohammad Ghassemi, Benjamin Moody, Peter Szolovits, Leo~Anthony Celi, and
  Roger~G Mark.
\newblock Data descriptor: Mimic-iii, a freely accessible critical care
  database (2016), 2016{\natexlab{b}}.

\bibitem[Khadanga et~al.(2019)Khadanga, Aggarwal, Joty, and
  Srivastava]{khadanga2019using}
Swaraj Khadanga, Karan Aggarwal, Shafiq Joty, and Jaideep Srivastava.
\newblock Using clinical notes with time series data for icu management.
\newblock \emph{arXiv preprint arXiv:1909.09702}, 2019.

\bibitem[Kingma and Ba(2014)]{kingma2014adam}
Diederik~P Kingma and Jimmy Ba.
\newblock Adam: A method for stochastic optimization.
\newblock \emph{arXiv preprint arXiv:1412.6980}, 2014.

\bibitem[Nori et~al.(2023)Nori, King, McKinney, Carignan, and
  Horvitz]{gpt4_medical-nori2023capabilities}
Harsha Nori, Nicholas King, Scott~Mayer McKinney, Dean Carignan, and Eric
  Horvitz.
\newblock Capabilities of gpt-4 on medical challenge problems, 2023.

\bibitem[Roy and Pan(2021)]{roy2021incorporating}
Arpita Roy and Shimei Pan.
\newblock Incorporating medical knowledge in bert for clinical relation
  extraction.
\newblock In \emph{Proceedings of the 2021 conference on empirical methods in
  natural language processing}, pages 5357--5366, 2021.

\bibitem[Soldaini and Goharian(2016)]{soldaini2016quickumls}
Luca Soldaini and Nazli Goharian.
\newblock Quickumls: a fast, unsupervised approach for medical concept
  extraction.
\newblock In \emph{MedIR workshop, sigir}, pages 1--4, 2016.

\bibitem[Umeicon et~al.(2023)Umeicon, Ware, juicy\_fish, and Squad]{icons}
Umeicon, Good Ware, juicy\_fish, and Vector Squad.
\newblock Flaticon, 2023.
\newblock URL \url{https://www.flaticon.com/}.

\bibitem[Veli{\v{c}}kovi{\'c} et~al.(2017)Veli{\v{c}}kovi{\'c}, Cucurull,
  Casanova, Romero, Lio, and Bengio]{velivckovic2017graph}
Petar Veli{\v{c}}kovi{\'c}, Guillem Cucurull, Arantxa Casanova, Adriana Romero,
  Pietro Lio, and Yoshua Bengio.
\newblock Graph attention networks.
\newblock \emph{arXiv preprint arXiv:1710.10903}, 2017.

\bibitem[Yang et~al.(2021)Yang, Kuang, and Xia]{yang2021multimodal}
Haiyang Yang, Li~Kuang, and FengQiang Xia.
\newblock Multimodal temporal-clinical note network for mortality prediction.
\newblock \emph{Journal of Biomedical Semantics}, 12\penalty0 (1):\penalty0
  1--14, 2021.

\bibitem[Yasunaga et~al.(2022)Yasunaga, Bosselut, Ren, Zhang, Manning, Liang,
  and Leskovec]{dragon-yasunaga2022deep}
Michihiro Yasunaga, Antoine Bosselut, Hongyu Ren, Xikun Zhang, Christopher~D
  Manning, Percy Liang, and Jure Leskovec.
\newblock Deep bidirectional language-knowledge graph pretraining, 2022.

\bibitem[Y{\`e}che et~al.(2021)Y{\`e}che, Kuznetsova, Zimmermann, H{\"u}ser,
  Lyu, Faltys, and R{\"a}tsch]{yeche2021hirid}
Hugo Y{\`e}che, Rita Kuznetsova, Marc Zimmermann, Matthias H{\"u}ser, Xinrui
  Lyu, Martin Faltys, and Gunnar R{\"a}tsch.
\newblock Hirid-icu-benchmark--a comprehensive machine learning benchmark on
  high-resolution icu data.
\newblock \emph{arXiv preprint arXiv:2111.08536}, 2021.

\bibitem[Zhang et~al.(2021)Zhang, Zeman, Tsiligkaridis, and
  Zitnik]{zhang2021graph}
Xiang Zhang, Marko Zeman, Theodoros Tsiligkaridis, and Marinka Zitnik.
\newblock Graph-guided network for irregularly sampled multivariate time
  series.
\newblock \emph{arXiv preprint arXiv:2110.05357}, 2021.

\bibitem[Zhang et~al.(2022)Zhang, Zhou, Song, Sui, Zhao, Jiang, and
  Yuan]{zhang2022pm2f2n}
Ying Zhang, Baohang Zhou, Kehui Song, Xuhui Sui, Guoqing Zhao, Ning Jiang, and
  Xiaojie Yuan.
\newblock Pm2f2n: Patient multi-view multi-modal feature fusion networks for
  clinical outcome prediction.
\newblock In \emph{Findings of the Association for Computational Linguistics:
  EMNLP 2022}, pages 1985--1994, 2022.

\end{thebibliography}

\clearpage
\appendix

\section{Dataset}
\label{sec:data}
\subsection{Pre-processing}
For the vital signs data, we follow the preprocessing steps outlined in~\cite{harutyunyan2019multitask}. For clinical text notes, we follow the preprocessing steps outlined in~\cite{husmann2022importance}. Essentially for all text notes that do not have CHARTTIME information, we set their CHARTTIME to the end of the respective CHARTDATE, remove all discharge summary notes, and then mask the last text note available. Post the filtering we exclude all patients without any text note similar to ~\cite{khadanga2019using}.     

\subsection{Task Statistics}
We provide the statistics for the test set on mortality and decompensation tasks.
\begin{table}[!h]
    \begin{minipage}{.25\textwidth}
      \caption{Decompensation}
      \centering
       \begin{tabular}{ll}
       \hline
        Label & \# labels \\ \hline
        0     & 511900           \\
        1     & 9403             \\
        Total & 521303           \\ \hline
       \end{tabular}
    \end{minipage}\begin{minipage}{.25\textwidth}
        \caption{Mortality}
        \centering
        \begin{tabular}{ll}
        \hline
        Label & \# labels \\ \hline
        0     & 2853           \\
        1     & 365             \\
        Total & 3218           \\ \hline
        \end{tabular}
    \end{minipage} 
\end{table}

\subsection{Clinical Notes}
\label{app:clinical-notes}
We provide some details on the clinical notes published in the MIMIC-III~\cite{johnson2016mimic} dataset in the \texttt{NOTEEVENTS} table.

In total, there are about 2 million individual text notes of 15 categories (e.g. \emph{Discharge summary}, \emph{Radiology}, \emph{Nursing}, \emph{Physician}, \emph{Nutrition}, \emph{General}, \emph{Pharmacy}, etc.). The median length of such a note is at $1,090$ characters and we observe a median of about 14 individual notes per patient (with 7 at the first quartile and 30 at the third quartile).

Each note is associated with a specific timestamp (\texttt{CHARTDATE} and \texttt{CHARTTIME}) during a single admission (\texttt{HADM\_ID}) of a given patient (\texttt{SUBJECT\_ID}). However, for a given patient admission we do not observe a note at every single time point on our resampled grid used during training. Some time points might have no clinical note associated with them, whereas others might have multiple, and they thus build an irregularly sampled time series of textual descriptions of the patient state. As such the set of clinical notes $\mathcal{N}_{t}^{i}$ at time $t$ given patient $i$ might be empty. In that case, the representation $\mathbf{n}_t^i$ is a 0-vector.

\section{Hyperparameters}
\label{sec:hyper}
We extract knowledge subgraphs $G_{t, KG}^{i}$ with a maximum of 30 nodes, use a 2-layered GNN, and set all latent space embedding dimensions to 64. We rely on the Adam optimizer~\citep{kingma2014adam} to train all the models with a learning rate of $1e^{-4}$. Details of other hyperparameters searched can be found in table \ref{tab:hyper}
\begin{table}[!ht]
\small
\caption{Hyperparameters for Experiments}
\begin{tabular}{ll}
\hline
\textbf{Parameters}         & \textbf{Values}                         \\ \hline
Batch Size         & (8, 16, 32, 64)                \\
Embedding Size     & (64)                           \\
GNN Depth          & (0,1,\textbf{2})                        \\
Graph Convolution  & (GCN, GAT, \textbf{GraphSAGE})          \\
Learning rate      & (\textbf{1e-4}, 1e-3)                   \\
Hidden Size (LSTM) & (100)                          \\
\# KG Nodes        & (0, 6, 18, \textbf{30}, 42, 54, 66, 78) \\
Nodes Aggregation  & (\textbf{sum}, mean, max)               \\
Epochs             & (20, 40)                       \\ \hline
\end{tabular}
\label{tab:hyper}
\end{table}

Parameters in bold indicate the choice that performed best. Batch size was varied across tasks so that the data could fit in the GPU memory. 20 epochs sufficed for mortality and decompensation tasks and 40 was used for phenotyping.

\section{Ablation Studies}
In this section, we show how the incremental addition of model components impacts performance.

\begin{table*}[!ht]
\small\centering
\caption{Mortality Performance Study under Different Model Components}
\begin{tabular}{lll}
\hline
\textbf{Model Component}                         & \textbf{AuPRC}        & \textbf{AuROC}        \\ \hline
LSTM (Vital Signs only)                 & 49.5 +/- 0.6 & 85.6 +/- 0.3 \\
LSTM-FT (Vital Signs only)              & 51.2 +/- 0.1 & 85.9 +/- 0.2 \\
LSTM-FT-GNN (Vital Signs only)          & 51.2 +/- 0.5 & 85.6 +/- 0.1 \\
LSTM (Vital Signs \& Text)              & 52.8 +/- 1.0 & 87.3 +/- 0.3 \\
LSTM-FT (Vital Signs \& Text)           & 55.2 +/- 0.9 & 87.7 +/- 0.1 \\
LSTM-FT-GNN (Vital Signs \& Text)       & 55.6 +/- 0.5 & 88.1 +/- 0.1 \\
LSTM-FT-GNN (Vital Signs \& Text \& KG) & \textbf{58.9 +/- 0.4} & \textbf{88.9 +/- 0.1} \\ \hline

\end{tabular}
\label{ihm}
\end{table*}

Table \ref{ihm} shows the performance variation on the mortality task when different model components/features are added incrementally. As seen in the case of Vital Signs-only data addition of the feature tokenizer module improves the performance, however further addition of the graph conv layer does not help much. With Vital Signs and Text data combined one can see a significant boost in performance due to the addition of the FT module, the addition of the graph conv layer does not improve overall performance. Finally, incorporating prior medical knowledge in the form of KG also helps and enhances performance.

\section{Interpretability Results}

\subsection{Analysis of Patient ID: 16236 (3rd Episode)}

\begin{figure*}[h!]
\centering
\includegraphics[width=\linewidth]{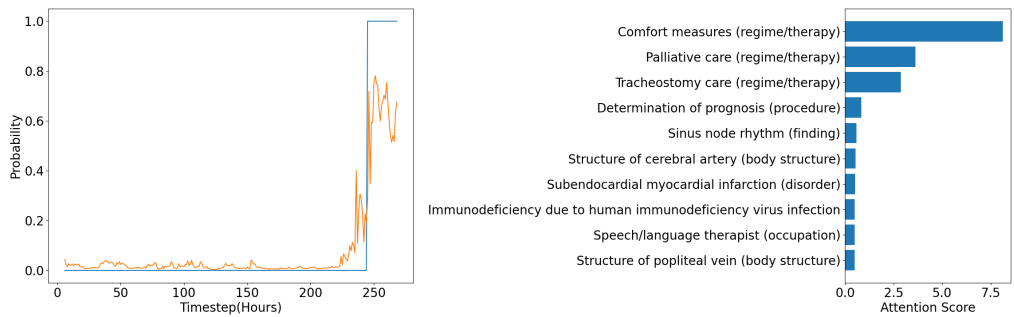}
\caption{Decompensation Analysis of Patient Id - 16236 : (left) \textit{orange} line shows the model's prediction and \textit{blue} line depicts the ground-truth label; (right) Attention Scores of KG nodes}
\label{fig:interpret2}
\end{figure*}

\begin{figure*}[h!]
\centering
\hspace*{1cm}\includegraphics[width=\linewidth]{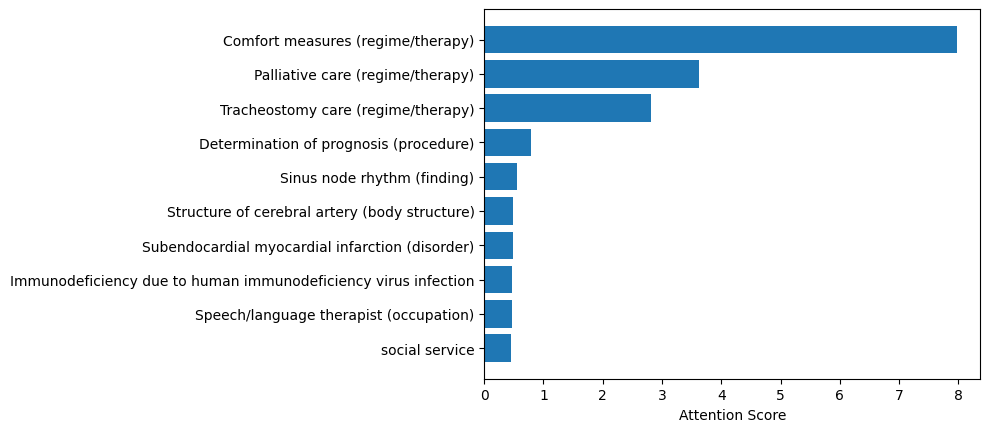}
\caption{HeatMap(s) at timestep 235}
\label{fig:interpret2b}
\end{figure*}

Figure \ref{fig:interpret2} gives another illustrative example to show the model's behavior on the decompensation task. In this example, different medical concepts related to therapies like \emph{Comfort Measures}, \emph{Palliative Care}, and \emph{Tracheostomy Care} form the most significant knowledge graph nodes. Figure~\ref{fig:interpret2} shows heatmaps at timestep 235. On the left heatmap, the 32nd entry has high attention and on inspection, it was the \emph{Palliative care} knowledge graph node similar to the previous example.

\subsection{Analysis of Patient ID: 18129 (1st Episode)}
\begin{figure*}[h!]
\centering
\includegraphics[width=\linewidth]{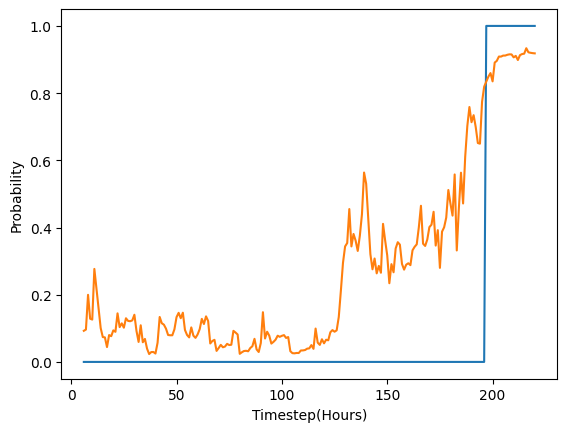}
\caption{Decompensation Analysis of Patient Id - 18129 : (left) \textit{orange} line shows the model's prediction and \textit{blue} line depicts the ground-truth label; (right) Attention Scores of KG nodes}
\label{fig:interpret3}
\end{figure*}

\begin{figure*}[h!]
\centering
\hspace*{1cm}\includegraphics[width=\linewidth]{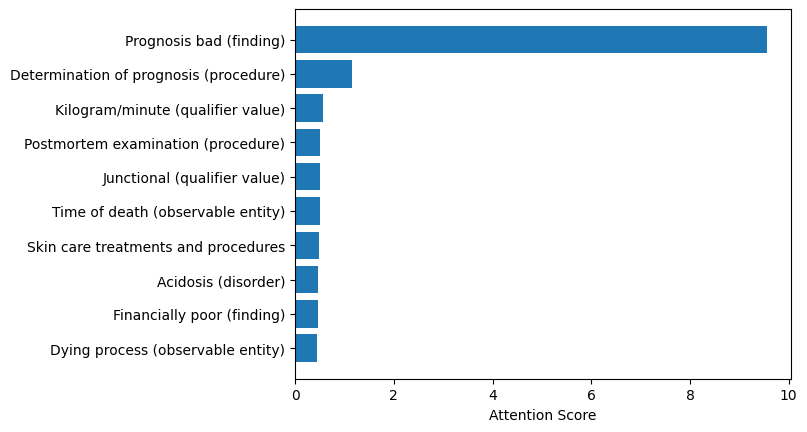}
\caption{HeatMap(s) at timestep 148}
\label{fig:interpret3b}
\end{figure*}

Figure \ref{fig:interpret3} gives another illustrative example. In this case, the prognosis bad node has an extremely high attention score. A bad prognosis means that the patient has very little chance of recovery. From the left heatmap of figure \ref{fig:interpret3b}, the 27th entry has high attention, and on inspection, to no surprise, the node was prognosis bad.

\end{document}